\documentclass[11pt,a4paper]{article}
\usepackage[hyperref]{naaclhlt2018}
\usepackage{times}
\usepackage{latexsym}
\usepackage{amssymb}
\usepackage{url}
\usepackage{booktabs} 
\usepackage{mathtools}
\usepackage{amsmath}
\usepackage{graphicx}
\newcommand\tab[1][0.5cm]{\hspace*{#1}}
\usepackage{tabularx}
\usepackage[T1]{fontenc}
\usepackage{enumitem}
\usepackage{multirow}

\usepackage[utf8]{inputenc}
\usepackage[english]{babel}
\usepackage{color,soul}
\usepackage{subfigure}
\usepackage[T1]{fontenc}

\aclfinalcopy 
\def\aclpaperid{184} 

\newcommand{\Ni}{({\em i})~}
\newcommand{\Nii}{({\em ii})~}
\newcommand{\Niii}{({\em iii})~}

\title{Automatic Stance Detection Using End-to-End Memory Networks}

\author{
	Mitra Mohtarami$^1$, Ramy Baly$^1$, James Glass$^1$, Preslav Nakov$^2$ \\
    {\bf Llu\'is M\`arquez$^3$\thanks{\hspace{.3em} This work was carried out when the authors were scientists at the Qatar Computing Research Institute, HBKU.}, Alessandro Moschitti$^{3\ast}$}\\
    $^1$MIT Computer Science and Artificial Intelligence Laboratory, Cambridge, MA, USA\\
    $^2$Qatar Computing Research Institute, HBKU, Doha, Qatar; $^3$Amazon\\
{\tt \{mitra,baly,glass\}@csail.mit.edu} \\
{\tt pnakov@hbku.edu.qa; \{lluismv,amosch\}@amazon.com} }
 
\date{}

\begin{document}
\maketitle

\begin{abstract}

We present a novel end-to-end memory network for stance detection, which jointly
\Ni~predicts whether a document \emph{agrees}, \emph{disagrees}, \emph{discusses} or is \emph{unrelated} with respect to a given target claim, and also
\Nii~extracts snippets of evidence for that prediction.  The network operates at the paragraph level and integrates convolutional and recurrent neural networks, as well as a similarity matrix as part of the overall architecture.
The experimental evaluation on the Fake News Challenge dataset shows state-of-the-art performance.
\end{abstract}

\section{Introduction}
\label{sec:introduction}

Recently, an unprecedented amount of false information has been flooding the Internet with aims ranging from affecting individual people's beliefs and decisions \cite{mihaylov-georgiev-nakov:2015:CoNLL,mihaylov-EtAl:2015:RANLP2015,ACL2016:trolls} to influencing major events such as political elections~\cite{Vosoughi1146}.
Consequently, manual fact checking has emerged with the promise to support accurate and unbiased analysis of public statements.

As manual fact checking is a very tedious task, automatic fact checking has been proposed as an alternative.
This is often broken into intermediate steps in order to alleviate the task complexity.
One such step is \emph{stance detection}, which is also useful for human experts as a stand-alone task.
The aim is to identify the relative perspective of a piece of text with respect to a claim, typically modeled using labels such as \emph{agree}, \emph{disagree}, \emph{discuss}, and \emph{unrelated}.
Figure~\ref{tbl:stance_example} shows some examples.

\begin{figure}[t]
\centering
\scalebox{0.75}{
\begin{tabular}{p{0.5in}|p{3in}}
\toprule
\multicolumn{2}{p{3.5in}}{{\bf Claim:} Robert Plant Ripped up \$800M Led Zeppelin Reunion Contract.}\\
\midrule
{\bf Stance} & {\bf Snippet}\\
agree & Led Zeppelin's Robert Plant turned down \pounds500m to reform supergroup...\\
disagree & Robert Plant's publicist has described as ``rubbish'' a Daily Mirror report that he rejected a \pounds500m Led Zeppelin reunion...\\
discuss & Robert Plant reportedly tore up an \$800 million Led Zeppelin reunion deal...\\
un-related & Richard Branson's Virgin Galactic is set to launch SpaceShipTwo today...\\
\bottomrule
\end{tabular}}
\caption{\label{tbl:stance_example}Examples of snippets of text and their stances with respect to a given claim.}
\end{figure}

Here, we address the problem using a novel model based on end-to-end memory networks~\cite{NIPS2015_5846}, which incorporates convolutional and recurrent neural networks, as well as a similarity matrix. 

\noindent Our model jointly addresses the problems of predicting the stance of a text document with respect to a given claim, and of extracting relevant text snippets as support for the prediction of the model. 
We further introduce a similarity matrix, which we use at inference time in order to improve the extraction of relevant snippets. 

The experimental results on the Fake News Challenge benchmark dataset show that our model, which is very feature-light, performs similarly to the state of the art, which is achieved by more complex systems.
Our contributions can be summarized as follows:
\Ni~We apply a novel memory network model enhanced with CNN and LSTM networks for stance detection.
\Nii~We further propose a novel extension of the general architecture based on a similarity-based matrix, which we use at inference time, and we show that this extension offers sizable performance gains.
\Niii~Finally, we show that our model is capable of extracting meaningful snippets from the input text document, which is useful not only for stance detection, but more importantly can be useful for human experts who need to decide on the factuality of a given claim.

\section{Model}

\label{sec:memNN_stance}
Long-term memory is necessary 
in order to determine the stance of a long document with respect to a claim, as
relevant parts of a document ---paragraphs or text snippets--- can indicate the perspective of a document with respect to a claim.
Memory networks were designed to remember past information~\cite{NIPS2015_5846} and they can be particularly well-suited for stance detection since they can use a variety of inference strategies alongside their memory component. 

In this section, we present a novel memory network (MN) for stance detection. It contains a new inference component that incorporates a similarity matrix
to extract, with better accuracy, textual snippets 
that are relevant to the input claims.

\subsection{Overview of the network}
A memory network is a 
5-tuple $\{M, I, G, O, R\}$, where the \emph{memory} $M$ is a sequence of objects or representations, the \emph{input} $I$ is a component that maps the input to its representation, the \emph{generalization} component $G$ \cite{NIPS2015_5846} updates the memory with respect to new input, the \emph{output} $O$ generates an output for each new input and the current memory state, and finally, the \emph{response} $R$ converts the output into a desired response format, e.g., a textual response or an action. These components can potentially use many different machine learning models.

\begin{figure*}[t]
\centering
\includegraphics[width=0.76\paperwidth]{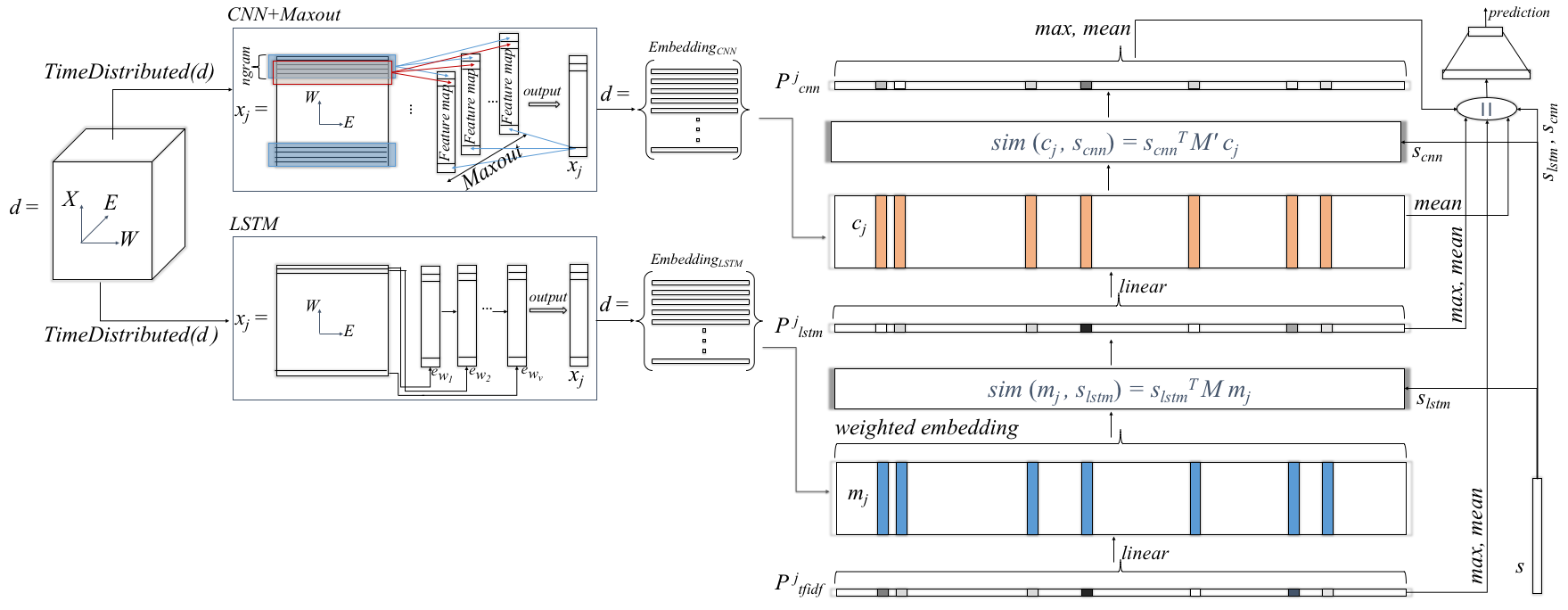}
\vspace{-.7em}
\caption{The architecture of our Memory Network model for stance detection.}
\label{fig:memNN}
\end{figure*}

Our new memory network for stance detection is a 6-tuple $\{M, I, F, G, O, R\}$, where $F$ represents the new \emph{inference} component. It takes an input document $d$ as evidence and a textual statement $s$ as a claim and converts them into their corresponding representations in the input  $I$. Then, it passes them to the memory  $M$. Next, the relevant parts of the input are identified in $F$, and afterwards they are used by $G$ to update the memory. 
Finally, $O$ generates an output from the updated memory, and converts it to a desired response format with $R$. The network architecture is depicted in Figure~\ref{fig:memNN}. We describe the components below.

\begin{table}
\small
\centering
\scalebox{0.9}{
\begin{tabularx}{1.15\linewidth}{|lX|}
\hline
1 & {\bf Inputs:}\\
2 & (1) A document ($d$) as a set of evidence ($x_j$) \\
3 & (2) A textual statement containing a claim ($s$) \\ 
[2mm]
4 & {\bf Outputs:}\\
5 & (1) predicting the relative perspective (or stance) of a pair of $(d,s)$ to a claim as \textit{agree}, \textit{disagree}, \textit{discuss} and \textit{unrelated}. \\
6 & \textit{Inference outputs:}\\
7 & (2) Top K evidence $x_j$ with their similarity scores \\
8 & (3) Top K snippets of $x_j$ with their similarity scores \\
[4mm]

9 & {\bf Memory Network Model:}\\
10 & \textit{1. Input memory representation (I):}\\
11 & \tab $d \rightarrow  ({X}, {W}, {E})$ \\ 
12 & \tab $(X, W, E) \xrightarrow{TimeDistributed(LSTM)} \{m_1,...,m_n\}$ \\
13 & \tab $(X, W, E) \xrightarrow{TimeDistributed(CNN)} \{c_1,..,c_n\}$ \\[1mm]
14 & \tab $ s \xrightarrow{LSTM, CNN} s_{\textit{lstm}}, s_{\textit{cnn}} $ \\
[2mm]

15 & \textit{2. Memory (M), updating memory (G) and inference (F):}\\
16 & \tab $m_j = m_j\odot P^{j}_{\textit{tfidf}}, \forall{j} $ \\
17 & \tab $P^{j}_{\textit{lstm}} = {s_{\textit{lstm}}}^\intercal\times\mathbf{M}\times m_j, \forall{j} $ \\
18 & \tab $c_j =  c_j\odot P^{j}_{\textit{lstm}}, \forall{j} $ \\
19 & \tab $P^{j}_{\textit{cnn}}=  {s_{\textit{cnn}}}^\intercal\times\mathbf{M'}\times c_j, \forall{j} $ \\
[2mm]

20 & \textit{3. Output memory representation (O):} $ o = \Big [ \textrm{mean}(\{c_j\}); $\\
21 & \tab $ \big [ \max(\{P^{j}_{\textit{cnn}}\});\textrm{mean}(\{P^{j}_{\textit{cnn}}\}) \big ]; \big [ \max(\{P^{j}_{\textit{lstm}}\})$; $ \tab \textrm{mean}(\{P^{j}_{\textit{lstm}}\}) \big ]; \big [ \max(\{P^{j}_{\textit{tfidf}}\}); \textrm{mean}(\{P^{j}_{\textit{tfidf}}\})  \big ] \Big ] $ \\
[2mm]

22 & \textit{4. Generating the final prediction (R):}\\
23 & \tab $[ o; s_{\textit{lstm}}; s_{\textit{cnn}}] \xrightarrow{MLP} \delta $\\
[2mm]

24 & \textit{5. Inference (F) outputs:}\\
25 & \tab $ P^{j}_{\textit{cnn}} \longrightarrow \{\emph{a set of evidences}\} + \{\textit{similarity scores}\} $ \\
26 & \tab $ M' \longrightarrow \{\emph{snippets}\} + \{\textit{similarity scores}\} $ \\
\hline
\end{tabularx}}
\caption{Summary of our Memory Network algorithm for stance detection.}\label{tbl:MemAlg}
\end{table}

\subsection{Input Representation Component}\label{subsec:input}

The input to the stance detection algorithm is a document $d$ and a textual statement $s$ as a claim: see lines 2 and 3 in Table~\ref{tbl:MemAlg}. Each $d$ is segmented into  paragraphs $x_j$ of varied lengths, where each $x_j$ is considered as a potential piece of evidence for stance detection. 

\noindent Indeed, a paragraph usually 
represents a coherent argument, unified under one or more inter-related topics. 
The input component in our model converts each $d$ into a set of potential pieces of evidence in a three-dimensional (3D) tensor space as shown below (see line 11 in Table~\ref{tbl:MemAlg}):
\begin{equation}\label{eq:input_d}
d = ({X}, {W}, {E})
\end{equation}
where ${X} = \{x_1, ..., x_n\}$ is a set of paragraphs considered as potential pieces of evidence, such that 
each $x_j$ is represented by a set of words ${W}=\{w_1, ..., w_v\}$---global vocabulary of size $v$---and a set of neural representations ${E}=\{e_1, ..., e_v\}$ for words in ${W}$. This 3D space is illustrated as a cube in Figure~\ref{fig:memNN}.

Each $x_j$ is encoded from the 3D space into a semantic representation at the input component using a 
Long Short-Term Memory (LSTM) network.
The lower left component in Figure~\ref{fig:memNN} shows our LSTM network, which operates on our input as follows (see also line 12 in Table~\ref{tbl:MemAlg}):
\begin{equation}\label{eq:LSTM}
({X}, {W}, {E}) \xrightarrow{TimeDistributed(LSTM)} \{m_1,...,m_n\}
\end{equation}

\noindent where $m_j$ is the LSTM representation of  $x_j$, and {\it TimeDistributed()} indicates a wrapper that enables training the LSTM over all pieces of evidence by applying the same LSTM model to each time-step of a 3D input tensor, i.e., $({X}, {W}, {E})$.

While LSTM networks are designed to effectively capture and memorize their inputs~\cite{acl:TanSXZ16}, Convolutional Neural Networks (CNNs) emphasize the local interaction between the words in the input word sequence,
which is important for obtaining an effective representation. We use a CNN to encode each $x_j$ into its representation $c_j$ as shown in Equation \ref{eq:CNN} (see line 13 in Table~\ref{tbl:MemAlg}).
\begin{equation}\label{eq:CNN}
(X, W, E) \xrightarrow{TimeDistributed(CNN)} \{c_1,..,c_n\}
\end{equation}
The left-top of Figure~\ref{fig:memNN} shows that this representation is passed as a new input to the component $M$ of our memory network.

We keep track of the computed $n$-grams from the CNN, so that we can use them later in the inference and in the response components (see Sections~\ref{subsec:inference}~and~\ref{subsec:response}). For this purpose, we use a Maxout layer~\cite{Goodfellow:2013} to take the maximum across $k$ affine feature maps computed by the CNN, i.e., pooling across channels.
Previous work has investigated the combination of convolutional and recurrent representations,
which is then fed to the other network as input \cite{acl:TanSXZ16,cvpr:DonahueHGRVDS15,Zuo_2015_CVPR_Workshops,ICASSP:Sainath}. In contrast, we feed their individual outputs into our memory network separately, and let the network decide which representation helps the target task better. We show the effectiveness of this choice below.  

\noindent Similarly, we convert each input claim $s$ to its representation using the corresponding LSTM and CNN networks, as follows:
\begin{equation}\label{eq:s_i}
s \xrightarrow{LSTM, CNN} s_{\textit{lstm}}, s_{\textit{cnn}}
\end{equation}
where $s_{\textit{lstm}}$ and $s_\textit{cnn}$ are the representations of $s$ computed using $LSTM$ and $CNN$ networks, respectively. Note that these are separate networks with different parameters from those  used to encode the pieces of evidence.
 
Lines 10--14 of Table~\ref{tbl:MemAlg} describe the above steps in representing $I$ in our memory network. We encode each input document $d$ into a set of pieces of evidence $\{x_j\} \forall{j}$: it computes LSTM and CNN representations, $m_j$ and $c_j$, respectively, for each $x_j$, and LSTM and CNN representations, $s_{\textit{lstm}}$ and $s_{\textit{cnn}}$, for each claim $s$.

\subsection{Inference Component}\label{subsec:inference}

\begin{figure}[t]
\centering
\includegraphics[width=\linewidth]{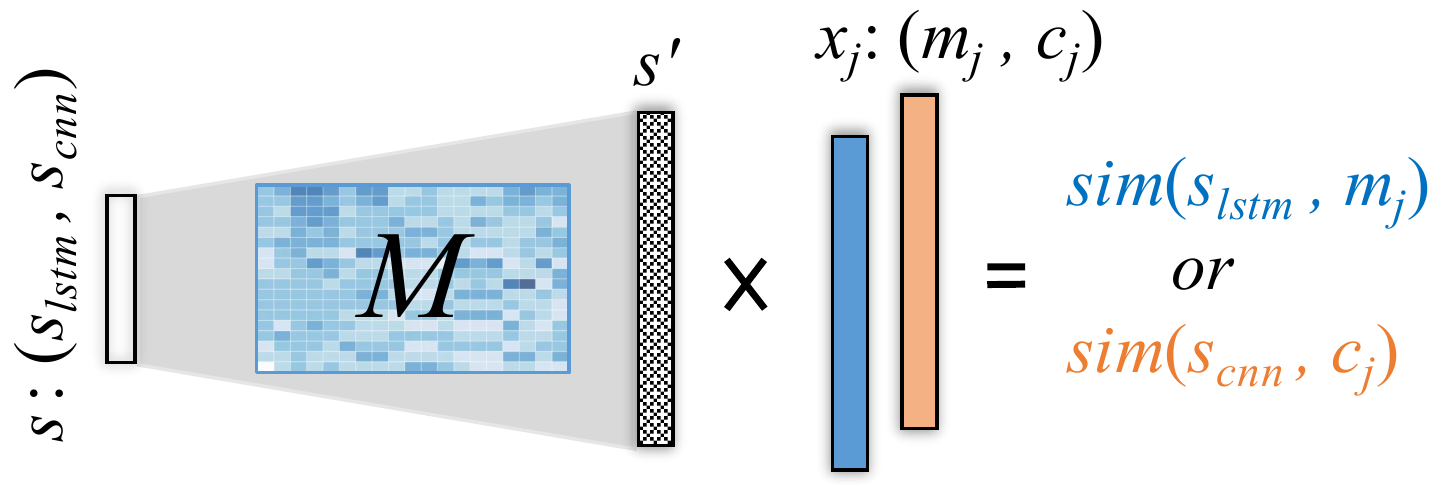}
\vspace{-1.3em}
\caption{Matching a claim $s$ and a piece of evidence $x_j$ using a similarity matrix $M$. Here, $s_{\textit{lstm}}$ and $s_{\textit{cnn}}$ are LSTM and CNN representations of $s$, whereas $m_j$ and $c_j$ are LSTM and CNN representations of $x_j$.}
\label{fig:simMatrix}
\vspace{-.7em}
\end{figure}

The resulting representations are used to compute semantic similarity between claims and pieces of evidence. 
We define the similarity $P^{j}_{\textit{lstm}}$
between $s$ and $x_j$ as follows (see also line 17 in Table~\ref{tbl:MemAlg}): 
\begin{equation}\label{eq:M_sim}
P^{j}_{\textit{lstm}} = {s_{\textit{lstm}}}^\intercal\times\mathbf{M}\times m_j, \forall{j}
\end{equation}
where $s_{\textit{lstm}}\in\mathbb{R}^q$ and $m_j\in\mathbb{R}^d$ are LSTM representations of $s$ and $x_j$, respectively, and $M\in\mathbb{R}^{q\times d}$ is a similarity matrix capturing their similarity. For this purpose, $M$ maps $s$ and $x_j$ into the same space as shown in Figure~\ref{fig:simMatrix}.  $M$ is a set of $q\times d$ parameters of the network, which are optimized during training.
         
In a similar fashion, we compute the similarity $P^{j}_{\textit{cnn}}$ between $x_j$ and $s$ using the CNN representations as follows (see line 19 of Table~\ref{tbl:MemAlg}): 
\begin{equation}\label{eq:M'_sim}
P^{j}_{\textit{cnn}}=  {s_{\textit{cnn}}}^\intercal\times\mathbf{M'}\times c_j, \forall{j}
\end{equation}
where $s_{\textit{cnn}}\in\mathbb{R}^{q'}$ and $c_j\in\mathbb{R}^{d'}$ are the representations of  $s$ and $x_j$ obtained with CNN, respectively. The similarity matrix $M'\in\mathbb{R}^{q'\times d'}$ is a set of $q'\times d'$ parameters of the network and is optimized during training. $P^{j}_{\textit{lstm}}$ and $P^{j}_{\textit{cnn}}$ indicate the claim-evidence similarity vectors computed based on the LSTM and on the CNN representations of $s$ and $x_j$, respectively.

The rationale behind using the similarity matrix is that in our memory network model, as Figure~\ref{fig:simMatrix} shows, we look for a transformation of the input claim $s$ such that $s' = M\times s$ in order to obtain the closest facts to the claim. 

In fact, the relevant parts of the input document with respect to the input claim can be captured at a different level, e.g., using $M'$ for the $n$-gram level or using the claim-evidence $P^{j}_{\textit{lstm}}$ or $P^{j}_{\textit{cnn}}, \forall{j}$ at the paragraph level. We note that \Ni  $P^j_{\textit{lstm}}$ uses LSTM to take the word order and long-length dependencies into account, and \Nii $P^j_{\textit{cnn}}$ exploits CNN to take $n$-grams and local dependencies into account, as explained in sections~\ref{subsec:input}~and~\ref{subsec:inference}.
Additionally, we compute another semantic similarity vector, $P^j_{\textit{tfidf}}$,  by applying a cosine similarity between the TF.IDF~\cite{sparck2004idf} representation of  $x_j$ and $s$.
This is particularly useful for stance detection as it can help detect the unrelated pieces of evidence.

\subsection{Memory and Generalization Components}\label{subsec:memory_update}

The information flow and updates in the memory is as follows: first, the representation vector $\{m_j\}\forall{j}$ is passed to the memory and updated using the claim-evidence similarity vector $\{P^{j}_{\textit{tfidf}}\}$:
\begin{equation}\label{eq:G_m}
m_j = m_j\odot P^{j}_{\textit{tfidf}}, \forall{j}
\end{equation}

The goal is to filter out most unrelated evidence. The updated $m_j$ in conjunction with $s_{\textit{lstm}}$ are used by the inference component--component $F$ to compute $\{P^j_{\textit{lstm}}\}$ as explained in Section~\ref{subsec:inference}. 

\noindent Then, $\{P^j_{\textit{lstm}}\}$ is used to update the new input set $\{c_j\}\forall{j}$ to the memory:
\begin{equation}\label{eq:G_c}
c_j =  c_j\odot P^{j}_{\textit{lstm}}, \forall{j}
\end{equation}

Finally, the updated $c_j$ in conjunction with $s_{\textit{cnn}}$ are used to compute $P^j_{\textit{cnn}}$  as explained in Sec.~\ref{subsec:inference}.

\subsection{Output Representation Component}\label{subsec:output}

In memory networks, the memory output depends on the final goal, which, in our case, is to detect the relative perspective of a document to a claim. For this purpose, we apply the following equation:
\begin{flalign}\label{eq:o}
\nonumber &o = \Big [ \textrm{mean}(\{c_{j}\}) ; & \\\nonumber & \big [ \max(\{P^{j}_{\textit{cnn}}\});\textrm{mean}(\{P^{j}_{\textit{cnn}}\})   \big ] ;
 \big [ \max(\{P^{j}_{\textit{lstm}}\}) ;& \\ &\textrm{mean}(\{P^{j}_{\textit{lstm}}\})    \big ] ; 
 \big [ \max(\{P^{j}_{\textit{tfidf}}\}) ; \textrm{mean}(\{P^{j}_{\textit{tfidf}}\})  \big ] \Big ]&
\end{flalign}
where $\textrm{mean}(\{c_{j}\})$ is the average vector of the $c_j$ representations.

Then, we compute the maximum and the average similarity between each piece of evidence and the claim using $P^j_{\textit{tfidf}}$, $P^j_{\textit{lstm}}$ and $P^j_{\textit{cnn}}$, which are computed for each evidence and claim in the inference component $F$. The maximum similarity identifies the part of document $x_j$ that is most similar to the claim, while the average similarity measures the overall similarity between the document and the claim.

\subsection{Response and Output Generation}\label{subsec:response}
This component computes the final stance of a document with respect to a claim. For this purpose, the concatenation of vectors $o$, $s_\textit{lstm}$ and $s_\textit{lstm}$, are fed into a Multi-Layer Perceptron (MLP), where a softmax predicts the stance of the document with respect to the claim, as shown below (see also lines $22$-$23$ in Table~\ref{tbl:MemAlg}):
\begin{equation}\label{eq:MLP}
[ o; s_{\textit{lstm}}; s_{\textit{cnn}}] \xrightarrow{MLP} \delta
\end{equation}
where $\delta$ is a softmax function. In addition to the resulting stance, we extract snippets from the input document that best indicates the perspective of the document with respect to the claim. For this purpose, we use $P^j_{\textit{lstm}}$, $P^j_{\textit{cnn}}$ and $M'$ as explained in Section~\ref{subsec:inference} (see also lines $24$-$26$ of Table~\ref{tbl:MemAlg}).

The overall model is shown in Figure~\ref{fig:memNN} and a summary of the model is presented in Table~\ref{tbl:MemAlg}. All model parameters, including those of \Ni CNN and LSTM in $I$, \Nii the similarity matrices $M$ and $M'$ in $F$, and \Niii the MLP in $R$, are jointly learned during the training process.

\section{Experiments and Evaluation}
\label{sec:experiments}

\subsection{Data}

We use the dataset provided by the Fake News Challenge,\footnote{Available at \url{www.fakenewschallenge.org}} where each example consists of a claim--document pair with the following possible relationship:
\emph{agree} (the document agrees with the claim),
\emph{disagree} (the document disagrees with the claim),
\emph{discuss} (the document discusses the same topic as the claim, but does not take a stance with respect to the claim),
\emph{unrelated} (the document discusses a different topic). 
The data includes a total of 75.4K claim-document pairs, which link 2.5K unique articles with 2.5K unique claims, i.e.,~each claim is associated with 29.8 articles on average. 

\subsection{Settings}
We use 100-dimensional word embeddings from GloVe~\cite{Pennington14glove:global}, which were pre-trained on two billion tweets. We use Adam
as an optimizer and categorical cross entropy as a loss function. 
We further use 100-dimensional units for the LSTM embeddings, and 100 feature maps with filter width of size 5 for the CNN. We consider the first $p$=9 paragraphs for each document, where $p$ is the median of the number of paragraphs.

We optimize the hyper-parameters of the models using the same validation dataset (20\% of the training data).
Finally, as the data is largely imbalanced towards the \emph{unrelated} class, during training we randomly select an equal number of instances from each class for each epoch. 

\subsection{Evaluation Measures}

We use the following evaluation measures:

{\bf Accuracy}: Number of correctly classified examples divided by the total number of examples. It is equivalent to micro-averaged F$_1$.

{\bf Macro-F1}: We calculate F$_1$ for each class, and then we average across all classes. 

{\bf Weighted Accuracy}: This is a weighted, two-level scoring scheme, which is applied to each test example. First, if the example is from the \emph{unrelated} class and the model correctly predicts it, the score is incremented by 0.25; otherwise, if the example is \emph{related} and the model predicts \emph{agree}, \emph{disagree}, or \emph{discuss}, the score is incremented by 0.25. Second, there is a further increment by 0.75 for each \emph{related} example if the model correctly predicts the correct label: \emph{agree}, \emph{disagree}, or \emph{discuss}. 

\noindent Finally, the score is normalized by dividing it by the total number of test examples.
The rationale behind this metric is that the binary \emph{related/unrelated} classification task is expected to be much easier, while also being arguably less relevant to fake news detection, than the actual stance detection task, which aims to further classify the relevant instances as \emph{agree}, \emph{disagree}, or \emph{discuss}. Therefore, the weighted accuracy metric gives more weight to the former distinction and less weight to the latter one.

\begin{table*}[t]
\centering
\scalebox{0.85}{
\begin{tabular}{p{4.5cm}@{ }cc|cccc}
\hline
  \bf Methods & \begin{tabular}{@{}l@{}}\bf{Total} \\ \bf{Parameters}\end{tabular} & \begin{tabular}{@{}l@{}}\bf{Trainable} \\ \bf{Parameters}\end{tabular} & \begin{tabular}{@{}l@{}}\bf{Weighted} \\ \bf{Accuracy}\end{tabular} & \bf Macro-F1 & \bf Accuracy \\ \hline
1. \ \ All-\emph{unrelated} & -- & -- & 39.37 & 20.96 & 72.20 \\
2. \ \ All-\emph{discuss} & -- & -- & 43.89 & 7.47 & 17.57 \\\hline
3. \ \ CNN & 2.7M & 188.7K & 40.66 & 24.44 & 41.53 \\
4. \ \ LSTM & 2.8M & 261.3K & 57.23 & 37.23 & 60.21\\
5. \ \ CNN+LSTM & 4.2M & 361.5K & 42.02 & 27.36 & 48.54\\
6. \ \ LSTM+CNN & 2.8M & 281.5K & 60.21 & 40.33 & 65.36\\
7. \ \ Gradient Boosting & -- & -- & 75.20 & 46.13 & 86.32 \\\hline
8. \ \ sMemNN (dotProduct) & 5.4M & 275.2K & 75.13 & 50.21 & 83.85\\
9. \ \ sMemNN & 5.5M & 377.5K & 78.97 & 56.75 & 87.27\\  
10.\ \ sMemNN (with TF) & 110M & 105M & \bf 81.23 & \bf 56.88 & \bf 88.57\\
\hline
\end{tabular}}
\caption{\label{tbl:results}Evaluation results on the test data.}
\end{table*}

\subsection{Baselines}

Given the imbalanced nature of our data, we use two baselines, in which we label all testing examples with the same label: (a)~\emph{unrelated} and (b)~\emph{discuss}. The former is the majority class baseline, which is a reasonable baseline for \emph{Accuracy} and \emph{macro-F$_1$}, while the latter is a potentially better baseline for \emph{Weighted Accuracy}.

We further use CNN and LSTM models, as well as combinations thereof, as baselines since they form components of our model, 
and also because they yield state-of-the-art results for text, image, and video classification~\cite{acl:TanSXZ16,cvpr:DonahueHGRVDS15,Zuo_2015_CVPR_Workshops,ICASSP:Sainath}. 

Finally, we include the official baseline from the challenge, which is a Gradient Boosting classifier with word and $n$-gram overlap features, as well as indicators for refutation and polarity.

\subsection{Our Models}

{\bf sMemNN}: This is our model presented in Figure~\ref{fig:memNN}. Note that unlike the CNN+LSTM and the LSTM+CNN baselines above, which feed the output of one network into the other one, the sMemNN model feeds the individual outputs of both the CNN and the LSTM networks into the memory network, and lets it decide how much to rely on each of them. This consideration also facilitates reasoning and explaining model predictions, as we will discuss in more detail below. 

{\bf sMemNN (dotProduct)}:  
This is a version of sMemNN, where the similarity matrices are replaced by the dot product between the representation of the claims and of the evidence. For this purpose, we first project the claim representation to a dense layer that has the same size as the representation of each piece of evidence, and then we compute the dot product between the resulting representation and the representation of the evidence.

{\bf sMemNN (with TF)}:  
Since our LSTM and CNN networks only use a limited number of starting paragraphs\footnote{Due to the long length of some documents, it is impractical to consider all paragraphs when training LSTM and CNN.} for an input document, we enrich our model with the BOW representation of documents and claims as well as their TF.IDF-based cosine similarity. These vectors are concatenated with the memory outputs (section~\ref{subsec:output}) and passed to the R component (section~\ref{subsec:response}) of sMemNN. We expect these BOW vectors to provide useful additional information.

\subsection{Results}

Table~\ref{tbl:results} reports the performance of all models on the test dataset. 
The All-\emph{unrelated} and the All-\emph{discuss} baselines perform poorly across the evaluation measures, except for All-\emph{unrelated}, which achieves high accuracy, which is due to \emph{unrelated} being by far the dominant class in the dataset. 

Next, we can see that LSTM consistently outperforms CNN across all evaluation measures. 
Although the larger number of parameters of the LSTM can play a role, we believe that its superiority comes from it being able to remember previously-observed relevant pieces of text.

Next, we see systematic improvements for the combinations of CNN and LSTM: CNN+LSTM is better than CNN alone, and LSTM+CNN is better than LSTM alone. Better performance is achieved by LSTM+CNN, that is, when claims and evidence are first processed by an LSTM network, and then fed into a CNN.

The Gradient Boosting model 
achieves sizable improvement over the above baseline neural models. However, we should note that these neural models do not use the rich hand-crafted features that were used in the Gradient Boosting model.

\noindent Row 9 shows the results for our memory network model (sMemNN), which consistently outperforms all other baseline models across all evaluation metrics, achieving 10.62 and 3.77 points of absolute improvement in terms of Macro-F1 and Weighted Accuracy, respectively, over the best baseline (Gradient Boosting). 
We believe that this is due to the memory network's capturing good text snippets. As we will see below, these snippets are also useful for explaining the model's predictions. 
Comparing row 9 to row 8, we can see the importance of our proposed similarity matrix: replacing that matrix by a simple dot product hurts the performance of the model considerably across all evaluation measures, thus lowering it to the level of the Gradient Boosting model.

Finally, row 10 shows the results for our memory network model enriched by a BOW representation. As we expected, it outperforms sMemNN, probably due to being able to capture useful information from paragraphs beyond the starting few.

To put the results of sMemNN in perspective, we should mention that the best system at the Fake News Challenge achieved a macro-F1 of 57.79, which is not significantly different from the performance of our full model at the 0.05 significance level (p-value=0.53). Yet, they have an ensemble combining the feature-rich Gradient Boosting system with neural networks. 

Further analysis of the output of the different systems (e.g., the confusion matrices) reveals the following general trends:
(\emph{i})~the \emph{unrelated} examples are easy to detect, and most models show high performance for this class, 
(\emph{ii})~the \emph{agree} and the \emph{disagree} examples are often mislabeled as \emph{discuss} by the baselines, and
(\emph{iii})~the \emph{disagree} examples are the most difficult ones for all models, probably because they represent by far the smallest class.

\begin{figure}
\centering
\includegraphics[width=0.9\linewidth]{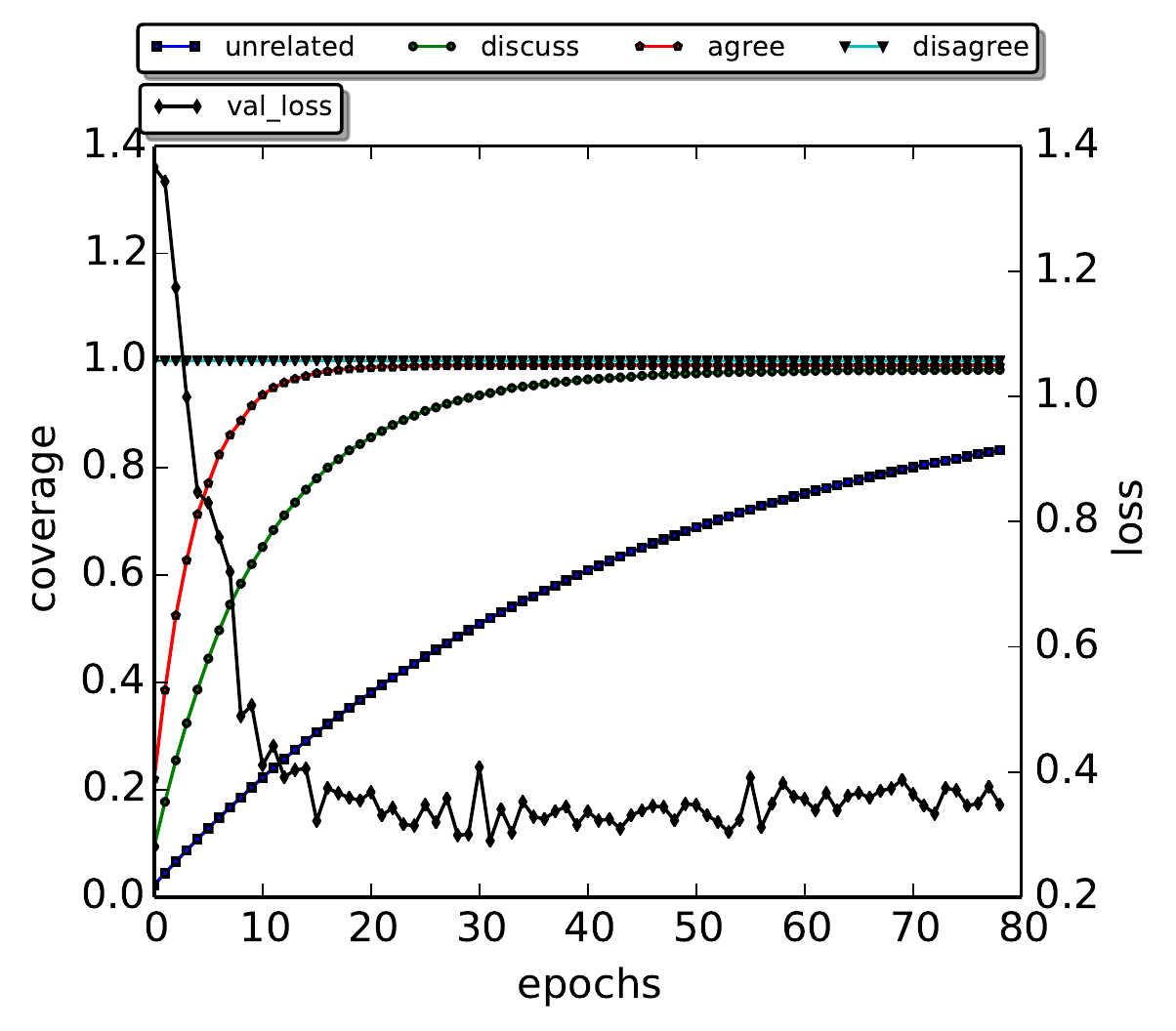}
\vspace{-1em}
\caption{Effect of data coverage. The $y$-axis shows the fraction of data observed during training (coverage), while the $x$-axis shows the loss during training.}
\label{fig:coverage}
\end{figure}

\begin{table*}
\small
\centering
\begin{tabular}{lcp{110mm}}
\hline
\multicolumn{3}{l}{\textbf{Claim 1:} man saved from bear attack - thanks to his justin bieber ringtone}\\
\hline
\textbf{Evidence Id} & \textbf{$\bf P^j_{cnn}$} & \textbf{Evidence Snippet}\\
\hline
2069-3 & 0.89 & ... fishing in the yakutia republic , russia , igor vorozhbitsyn is lucky to be alive after his justin bieber ringtone , \colorbox{cyan}{baby , scared off a bear that was attacking him}\textsuperscript{\bf{0.41}} ...\\ 
2069-7  & 1.0 & ... but as the bear clawed vorozhbitsyn ' s face and back his mobile phone rang , the ringtone selected was justin bieber ' s hit \colorbox{cyan}{song baby . rightly startled}\textsuperscript{\bf{1.00}} , \colorbox{cyan}{the bear retreated back into}\textsuperscript{\bf{0.39}} the forest ...\\
\hline
\multicolumn{3}{l}{\textbf{true label:} \textit{agree}; \textbf{predicted label:} \it{agree}}\\
\hline\\
\hline
\multicolumn{3}{l}{\textbf{Claim 2:} 50ft crustacean , dubbed crabzilla , photographed lurking beneath the waters in whitstable}\\
\hline
\textbf{Evidence Id} & \textbf{$\bf P^j_{cnn}$} & \textbf{Evidence Snippet}\\
\hline
24835-1 & 0.0046 & ... a marine \colorbox{cyan}{biologist has killed off claims}\textsuperscript{\bf{-0.0008}} \colorbox{cyan}{that a giant crab is}\textsuperscript{\bf{0.0033}} living on the kent coast - insisting the image is probably a \colorbox{cyan}{well - doctored hoax}\textsuperscript{\bf{0.0012}} ...\\ 
24835-7  & -0.0008 & ... i don ' t know what the currents are like around that harbour or what sort of they might produce in the sand , but i think it ' s more \colorbox{cyan}{conceivable that someone is playing}\textsuperscript{\bf{0.0007}} about with the photo ...\\
\hline
\multicolumn{3}{l}{\textbf{true label:} \textit{disagree}; \textbf{predicted label:} \it{disagree}}\\
\hline
\end{tabular}

\caption{Examples of highly ranked snippets of evidence for an input claim, which were automatically extracted by our inference component for claim-document pairs. The $P^j_{cnn}$ column and the values in the top-right corner of the highlighted snippets show the similarity between the claim and a piece of evidence, and between the claim and an evidence snippet, respectively.}
\label{tbl:snippets_inference}
\end{table*}

\section{Discussion}\label{sec:discussion}

\subsection{Training Data Coverage}
\label{dis:Coverage}

As discussed previously, we balance the data at each training iteration by randomly selecting $z$ instances from each of the four target classes,
where $z$ is the size of the class with the minimum number of training instances. In this experiment, we investigate what proportion of the training data got actually used when following our sampling procedure. For this purpose, at each training iteration, we report the proportion of the training instances from each class that were used so far, either at the current or at any of the previous iterations.

\noindent As Figure~\ref{fig:coverage} shows, our random data sampling procedure eventually used almost all training examples. Since the \emph{disagree} class was the smallest, its examples remained fully covered throughout the process. Moreover, almost all other related examples, i.e., \emph{agree} and \emph{discuss}, were observed during training, as well as a large fraction of the dominating \emph{unrelated} examples. 
Note that the model achieved its best (lowest) loss on the validation dataset at iteration 31, when almost all \emph{related} instances had already been observed. This happened while the corresponding fraction for the \emph{unrelated} pairs was around 50\%, i.e., a considerable number of the \emph{unrelated} instances were not really needed.

\subsection{Explainability}
\label{dis:inference}

A major advantage of our model, compared to the baselines and to most related work, is that it can explain its predictions:
as we explained in section~\ref{subsec:inference}, our inference component predicts the similarity between each piece of evidence $x_j$ and the claim $s$ at the $n$-grams-level using the claim-evidence similarity vector $P^j_{cnn}$.

Table~\ref{tbl:snippets_inference} shows examples of two claims and the snippets extracted as evidence. Column $P^j_{cnn}$ shows the overall similarity between the evidence and the corresponding claim as computed by the inference component of our model. The highlighted texts are snippets with the highest similarity (the value is shown next to each snippet) to the claim as extracted by the inference component. 

\noindent Note that the snippets are of fixed length, namely 5-grams, but in case of consecutive $n$-grams with similar scores, we combine them into a single snippet and we report the average value, e.g.,~see the snippet for evidence 2069-3. 
The lower half of Table~\ref{tbl:snippets_inference} shows an example where the similarity values associated with the snippets are either too small or negative, e.g., see the value for \emph{biologist has killed off claims}. 
In all cases, the model could accurately predict the stance of these pieces of evidence with respect to the corresponding claims. 

\begin{figure*}[t]
\centering
\subfigure[$n$-grams]{\label{fig:ngram}\includegraphics[scale=0.53]{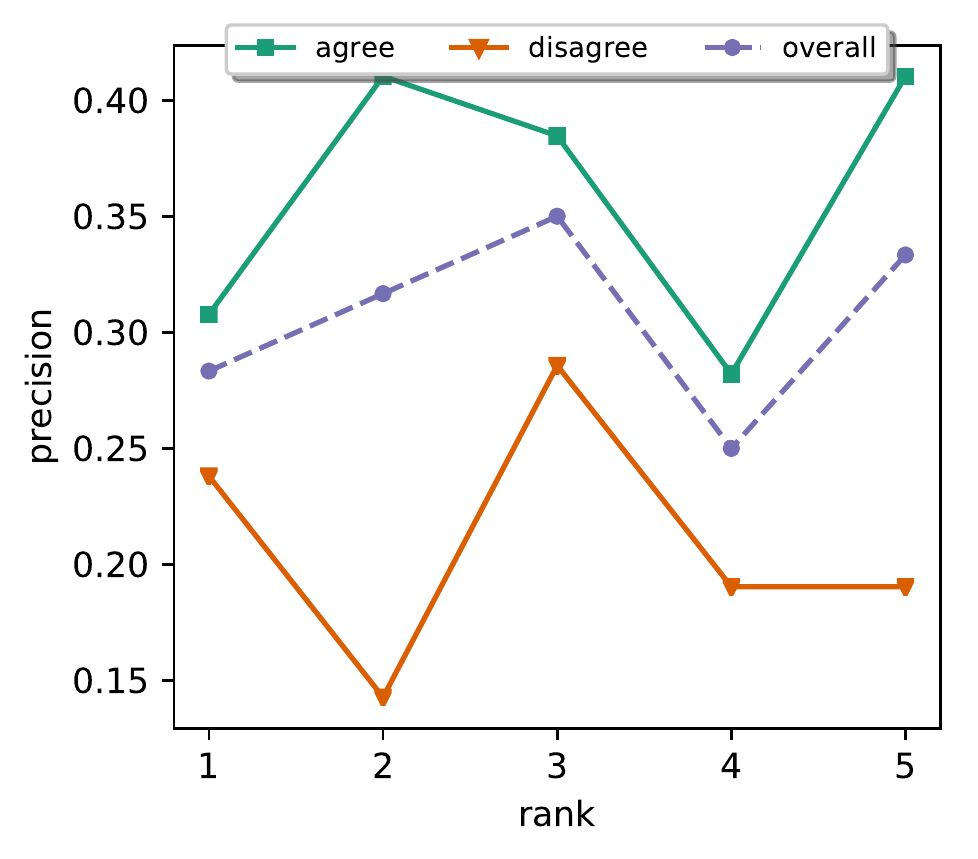}}
\subfigure[consecutive $n$-grams]{\label{fig:consecutive}\includegraphics[scale=0.53]{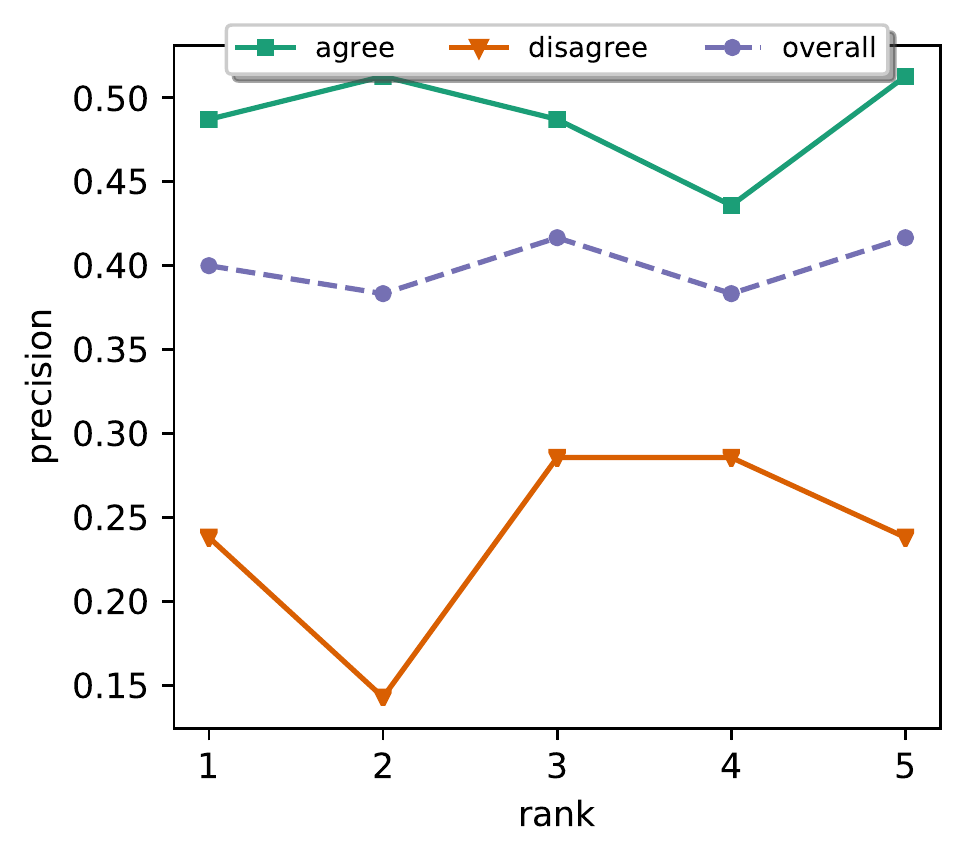}}
\subfigure[sentences]{\label{fig:sentence}\includegraphics[scale=0.53]{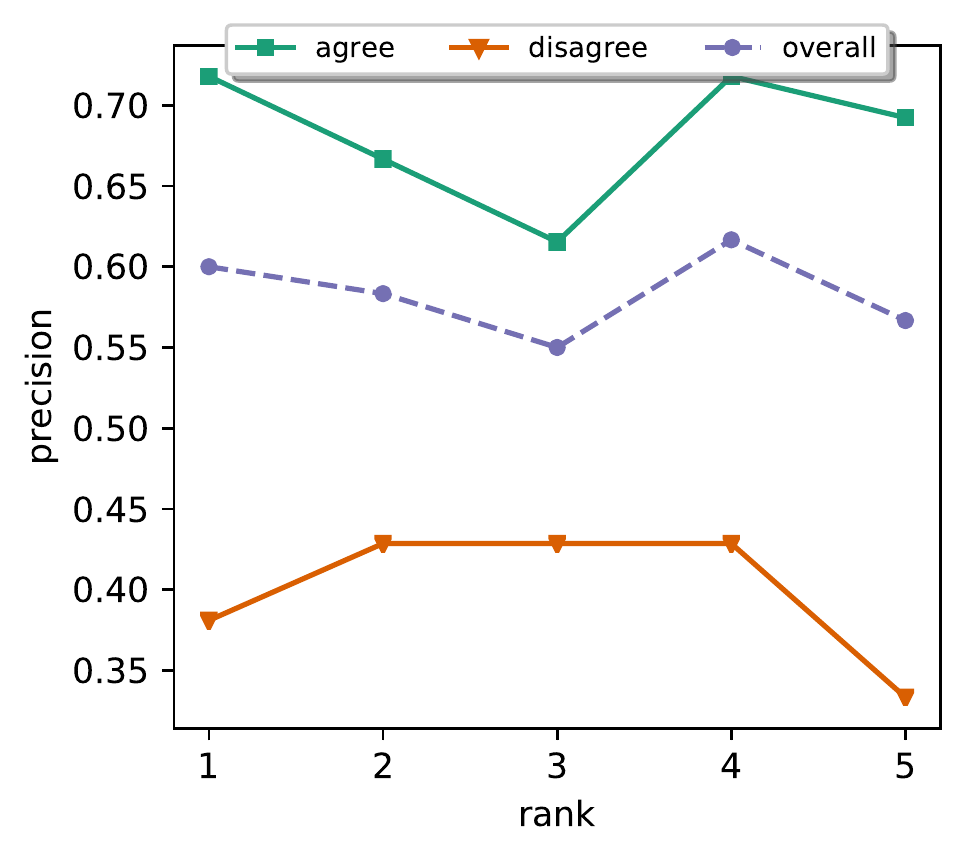}}
\caption{Prediction explainability. Sub-figures (a)-(c) show the precision of our model explaining its prediction when the pieces of evidence are (a)~fixed-length $n$-grams ($n=5$), (b)~combinations of several consecutive $n$-grams with similar scores, or (c)~the entire sentence, if it includes at least one extracted $n$-gram snippet.}
\label{fig:evidence_pre}
\end{figure*}

Next, we conducted an experiment to quantify the performance of our memory network at explaining its predictions: we randomly sampled 100 agree/disagree claim-document examples from our gold data, and we manually evaluated the top five pieces of evidence that our model provided. In 76 cases, the model correctly classified the agree/disagree examples, and provided arguably adequate snippets.

Figure~\ref{fig:ngram} shows the performance of our model at explaining its predictions when each supporting/opposing piece of evidence is an $n$-gram snippet of fixed length ($n=5$) for the \emph{agree} and the \emph{disagree} classes, and their combinations at the top-$k$ ranks, $k=\{1,\dots,5\}$. 
It achieved precision of 0.28, 0.32, 0.35, 0.25, and 0.33 at ranks 1--5. Moreover, we found that it could accurately identify, as part of the identified $n$-grams, key phrases such as \emph{officials declared the video}, \emph{according to previous reports}, \emph{believed will come}, \emph{president in his tweets} as supporting pieces of evidence, and \emph{proved a hoax}, \emph{shot down a cnn report}, \emph{would be skeptical} as opposing pieces of evidence. 

Note that the above low precision is mainly due to the unsupervised nature of this task as no gold snippets supporting the document's stance are available for training in the FNC dataset.\footnote{Some other recent datasets, to be presented at this same HLT-NAACL'2018 conference, do have such gold evidence annotations \cite{baly2018integrating,thorne2018fever}.} Furthermore, our evaluation setup was at the $n$-gram level in Figure~\ref{fig:ngram}. However, if we conduct a more coarse-grained evaluation where we combine consecutive $n$-grams with similar scores into a single snippet, the precision for these new snippets improves to 0.4, 0.38, 0.42, 0.38, and 0.42 at ranks~1--5, as Figure~\ref{fig:consecutive} shows. If we further extend the evaluation to the sentence level, the precision jumps to 0.6, 0.58, 0.55, 0.62, and 0.57 at ranks 1--5, as we can see on Figure~\ref{fig:sentence}.

\section{Related Work}
\label{sec:related_work}

While stance detection is an interesting task in its own right, e.g., for media monitoring, it is also an important component for fact checking and veracity inference.\footnote{Yet, stance detection and fact checking are typically supported by separate datasets. Two notable upcoming exceptions, both appearing in this HLT-NAACL'2018, are \cite{thorne2018fever} for English and \cite{baly2018integrating} for Arabic.}
Automatic fact checking was envisioned by \newcite{vlachos2014fact} as a multi-step process that
(\emph{i})~identifies check-worthy statements~\cite{hassan2015detecting,gencheva2017context,NAACL2018:claimrank},
(\emph{ii})~generates questions to be asked about these statements~\cite{karadzhov2017fully},
(\emph{iii})~
retrieves relevant information to create a knowledge base~\cite{shiralkar2017finding}, 
and
(\emph{iv})~infers the veracity of these statements, e.g., using text analysis~\cite{banerjee-han:2009:NAACLHLT09-Short,Castillo:2011:ICT:1963405.1963500,rashkin2017truth}
or information from external sources~\cite{karadzhov2017fully,Popat:2017:TLE:3041021.3055133}.

There have been some nuances in the way researchers have defined the stance detection task.
SemEval-2016 Task~6~\cite{mohammad2016semeval} targets stances with respect to some target proposition, e.g., entities, concepts or events, as \emph{in-favor}, \emph{against}, or \emph{neither}.
The winning model in the task was based on transfer learning: a Recurrent Neural Network trained on a large Twitter corpus was used to predict task-relevant hashtags and to initialize a second recurrent neural network trained on the provided dataset for stance prediction~\cite{zarrella2016mitre}.
Subsequently, \newcite{zubiaga2016stance} detected the stance of tweets toward rumors and hot topics using linear-chain conditional random fields (CRFs) and tree CRFs that analyze tweets based on their position in tree-like conversational threads. 

Most commonly, stance detection is defined 
with respect to a \emph{claim}, e.g., as in the 2017 Fake News
Challenge. 
The best system was an ensemble of gradient-boosted decision trees with rich features and CNNs~\cite{baird2017talos}.
The second system was a multi-layer neural network with similarity features, word $n$-grams, and latent semantic analysis \cite{hanselowski2017athene}.
The third one was a neural network with similarity features \cite{riedel2017simple}.

Unlike the above work, we use a feature-light memory network that jointly infers the stance and highlights relevant snippets of evidence.

\section{Conclusion}\label{sec:conclusion}

We studied the problem of stance detection, which aims to predict whether a document supports, challenges, or just discusses a given claim.
The nature of the task clearly shows that, in order to go beyond simple matching between stance (short text) and evidence (longer text, e.g., an entire document), a machine learning model needs to focus on the relevant paragraphs of the evidence. Moreover, in order to understand whether a paragraph supports a claim, there is a need to refer to information available in other paragraphs. 
CNNs and LSTMs are not well-suited for this task as they cannot model complex dependencies such as semantic relationships with respect to entire previous paragraphs. In contrast, memory networks are exactly designed to remember previous information. However, given the large size of documents and paragraphs, basic memory networks do not handle well irrelevant and noisy information, which we confirmed in our experimental results.

Thus, we proposed a novel extension of the basic memory networks, which is based on a similarity matrix and a stance filtering component, which we apply at inference time, and we have shown that this extension offers sizable performance gains,
making memory networks competitive.
Moreover, our model can extract meaningful snippets from documents that can explain the factuality of a given claim.

In future work, we plan to extend the inference component to select an optimal set of explanations for each prediction, and to explain the model as a whole, not only at the instance level.

\section*{Acknowledgment}\label{sec:acknowledgment}
We would like to thank the members of the MIT Spoken Language Systems group and the anonymous reviewers for their helpful comments. 

This research was carried out in collaboration between the MIT Computer Science and Artificial Intelligence Laboratory (CSAIL) and the Qatar Computing Research Institute (QCRI), HBKU.

\bibliographystyle{acl_natbib}
\bibliography{references}
\end{document}